\documentclass[pdflatex,sn-basic,Numbered]{sn-jnl}

\usepackage{graphicx}
\usepackage{xcolor}
\usepackage{booktabs}
\usepackage{amsmath,amssymb,amsfonts}
\usepackage{manyfoot}

\title{DALight-3D: A Lightweight 3D U-Net for Brain Tumor Segmentation from Multi-Modal MRI}

\author[1]{\fnm{Nand Kumar} \sur{Mishra}}

\author[2]{\fnm{Dhruv} \sur{Mishra}}

\author[1]{\pfx{Dr} \fnm{Manu Pratap} \sur{Singh}}

\affil[1]{\orgdiv{Department of Computer Science}, \orgname{Dr.\ Bhimrao Ambedkar University}, \orgaddress{\city{Agra}, \country{India}}}

\affil[2]{\orgdiv{Department of Computer Science and Engineering}, \orgname{Shiv Nadar University}, \orgaddress{\city{Greater Noida}, \country{India}}}

\abstract{Automatic brain tumor segmentation from multi-modal MRI remains challenging because volumetric models often incur substantial computational cost. This paper presents \textbf{DALight-3D}, a compact 3D U-Net variant that combines depthwise separable 3D convolutions, identifier-conditioned normalization, cross-slice attention, and adaptive skip fusion. The method is evaluated on the Medical Segmentation Decathlon Task01\_BrainTumour benchmark under matched optimization settings against standard 3D U-Net, Attention U-Net, Residual 3D U-Net, and V-Net baselines. In the reported 50-epoch comparison, DALight-3D achieves a mean Dice of 0.727 with 2.22M parameters, compared with 0.710 Dice and 3.20M parameters for Residual 3D U-Net. Component-wise ablations show consistent performance degradation when SepConv, identifier-conditioned normalization, CSA, or SSFB is removed. These results indicate that DALight-3D offers a favorable accuracy--efficiency trade-off within the present benchmark setting.}


\begin{document}

\maketitle

\section{Introduction}
\label{sec:intro}

Accurate delineation of brain tumors from multi-modal magnetic resonance imaging (MRI) is a critical component of modern neuro-oncology workflows. Reliable segmentation supports treatment planning, surgical navigation, radiotherapy targeting, and longitudinal monitoring of disease progression or treatment response~\cite{menze2015brats}. In clinical practice, tumor regions are typically annotated into several substructures, including background (BG), necrotic and non-enhancing tumor core (NCR), peritumoral edema (ED), and enhancing tumor (ET). However, manual delineation of these regions is labor-intensive and subject to inter-observer variability, which has motivated extensive research into automated segmentation methods.

\begin{figure}[t]
\centering
\includegraphics[width=\linewidth]{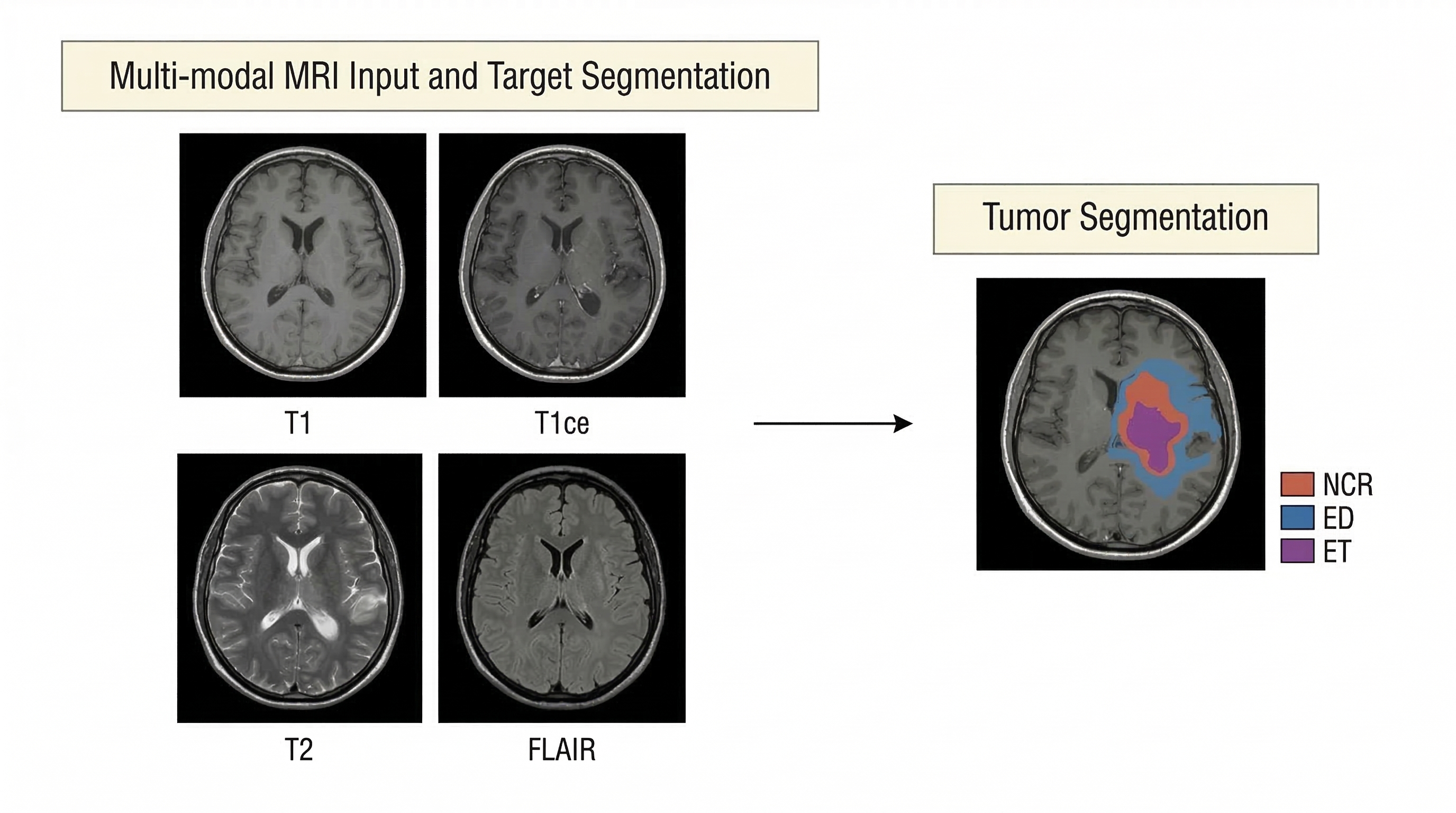}
\caption{Segmentation Setting \\ Left: the same lesion appears differently across T1, T1ce, T2, and FLAIR \\ Right: the target comprises NCR, ED, and ET subregions}
\label{fig:intro_multimodal}
\end{figure}

Deep learning has significantly advanced medical image segmentation. Encoder--decoder architectures, particularly U-Net~\cite{ronneberger2015unet}, established a widely adopted design paradigm in biomedical imaging by combining hierarchical feature extraction with skip connections that preserve spatial detail. Subsequent extensions such as 3D U-Net~\cite{cicek20163dunet} enabled direct volumetric analysis, substantially improving performance on MRI and CT volumes. Benchmark datasets and challenges, including BraTS~\cite{menze2015brats} and the Medical Segmentation Decathlon~\cite{simpson2019msd}, further accelerated progress. Numerous architectural refinements have since been proposed, incorporating residual connections~\cite{he2016resnet}, attention mechanisms~\cite{oktay2018attentionunet}, and deeper feature hierarchies to improve segmentation accuracy.

Figure~\ref{fig:intro_multimodal} summarizes the core learning problem addressed in this work: the model must exploit complementary multi-modal evidence to produce fine-grained voxel-wise delineation of heterogeneous tumor subregions. This is challenging because the regions of interest are not equally visible across modalities, and the final segmentation depends on how effectively these partially overlapping cues are fused.

Despite this progress, two practical limitations remain central to real-world deployment.

\begin{figure}[t]
\centering
\includegraphics[width=\linewidth]{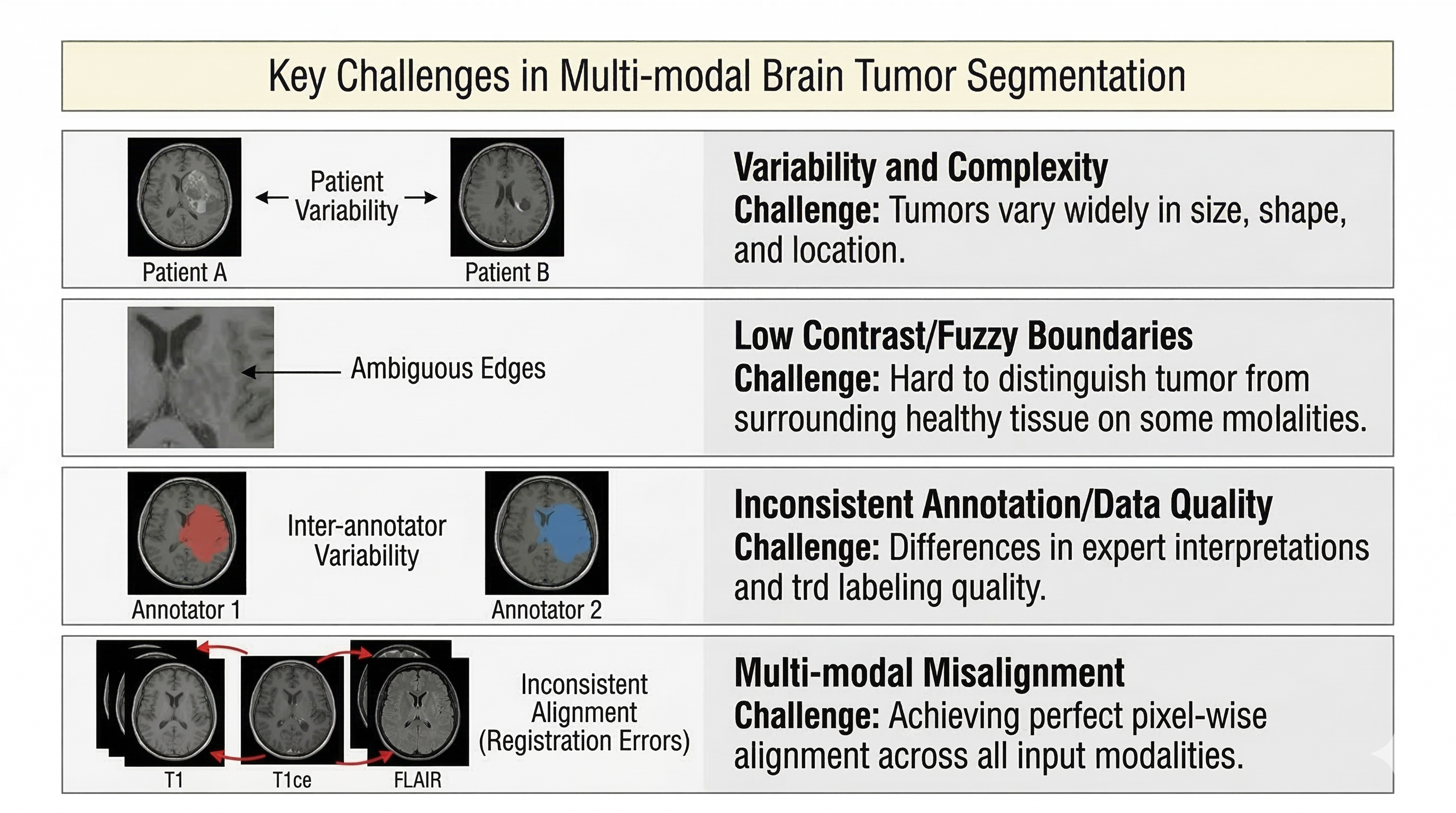}
\caption{Segmentation Challenges Variability in anatomy, tumor morphology, annotations, and modality alignment increases the difficulty of reliable voxel-wise prediction}
\label{fig:intro_challenges}
\end{figure}

As illustrated in Figure~\ref{fig:intro_challenges}, the difficulty of brain tumor segmentation is not limited to classification capacity alone. The model must remain reliable under patient-level variability, fuzzy lesion boundaries, annotation uncertainty, and cross-modality inconsistencies. These factors motivate architectures that are accurate, efficient, and context-aware under heterogeneous acquisition conditions.

\paragraph{Limitation 1: computational burden.}
High-performing volumetric CNNs often rely on dense $3\times3\times3$ convolutions and rapidly expanding channel widths, resulting in substantial parameter counts, memory requirements, and inference cost. In 3D, convolutional complexity grows as $O(C_{\mathrm{in}}C_{\mathrm{out}}k^3)$, making efficiency a first-order design constraint. Although lightweight operators such as depthwise separable convolutions have been highly effective in 2D vision~\cite{howard2017mobilenets}, they remain comparatively underexplored in compact 3D brain tumor segmentation models.

\paragraph{Limitation 2: acquisition variability.}
Another major challenge arises from variation in scanner hardware, acquisition protocols, and preprocessing pipelines. Standard normalization layers improve optimization stability but do not explicitly model acquisition-dependent feature statistics. Consequently, models trained on one distribution may remain sensitive to heterogeneity at deployment time.

\paragraph{Proposed approach.}
To address these issues, we propose \textbf{DALight-3D}, a compact volumetric segmentation framework that targets efficiency and stronger feature conditioning while preserving the encoder--decoder topology of U-Net. DALight-3D integrates four complementary components: (1) \emph{depthwise separable 3D convolutions} (SepConv) to reduce parameter and computational cost; (2) an \emph{identifier-conditioned normalization} module (ScannerAwareNorm); (3) \emph{cross-slice attention} (CSA) in deeper encoder stages to capture long-range inter-slice context at low cost; and (4) a \emph{Skip and Spatial Feature Blend} (SSFB) module that replaces naive skip concatenation with low-rank attention and channel gating for adaptive decoder--encoder fusion.

The key idea is to retain the representational strengths of 3D U-Net-style models while introducing targeted mechanisms for feature conditioning and computational efficiency. Rather than simply enlarging the backbone, DALight-3D improves feature extraction, normalization, and skip fusion under realistic deployment constraints.

\paragraph{Contributions.}
The main contributions of this work are as follows:
\begin{itemize}
\item We introduce DALight-3D, a lightweight 3D segmentation architecture that unifies separable volumetric convolutions, conditioned normalization, cross-slice attention, and adaptive skip fusion within a single framework.
\item We formulate a practical design strategy for jointly improving computational efficiency and identifier-conditioned feature calibration in multi-modal brain tumor segmentation.
\item We provide a reproducible implementation and evaluation pipeline, including parameter-aware comparison, ablation analysis, per-class metrics, confusion-matrix-based assessment, and calibration analysis.
\end{itemize}

The remainder of this paper is organized as follows. Section~\ref{sec:lit} reviews related work in brain tumor segmentation, efficient volumetric modeling, and robustness under acquisition variability. Section~\ref{sec:method} presents the proposed DALight-3D architecture. Section~\ref{sec:eval} describes the experimental protocol and quantitative evaluation. Section~\ref{sec:conclusion} concludes the paper.

\section{Literature Review}
\label{sec:lit}

Brain tumor segmentation from multi-modal MRI has evolved through a sequence of methodological shifts. Early classical pipelines relied on hand-crafted priors, clustering, deformable models, and intensity heuristics to separate tumor tissue from surrounding anatomy. These approaches were clinically interpretable but highly sensitive to scanner variability, noise, and tumor heterogeneity, motivating the need for more robust feature-learning methods~\cite{gordillo2013survey}.

\subsection{From Hand-Crafted Methods to CNN-Based Segmentation}

The first major transition came with convolutional neural networks, which replaced manually designed features with data-driven hierarchical representations. Early CNN-based methods for brain tumor segmentation demonstrated that learned features can outperform hand-crafted descriptors, especially when tumor appearance varies across patients~\cite{havaei2017brainseg}. Hybrid representation-learning strategies further showed that combining deep and complementary feature abstractions can improve segmentation and classification robustness in MRI~\cite{farajzadeh2023hybrid}. However, many early CNN pipelines remained patch-based or locally constrained, limiting their ability to integrate long-range volumetric context.

\subsection{The Rise of U-Net and Volumetric Encoder--Decoder Models}

U-Net~\cite{ronneberger2015unet} introduced the now-standard combination of hierarchical feature extraction and skip connections for dense prediction, and 3D U-Net~\cite{cicek20163dunet} extended this idea to volumetric MRI. V-Net~\cite{milletari2016vnet} further established fully volumetric learning with Dice-oriented optimization. These models significantly improved segmentation quality, but their success also exposed a trade-off: as 3D models became deeper and wider, their computational cost and memory footprint increased rapidly. Subsequent methods attempted to address this through cascaded anisotropic convolutions~\cite{wang2019cascaded}, multi-planar fusion~\cite{andermatt2018multiplanar}, and autoencoder-regularized volumetric segmentation~\cite{myronenko2019autoencoder}. Strong baselines such as No New-Net and nnU-Net showed that careful data processing, normalization, and optimization can be as important as architectural novelty~\cite{isensee2018nonewnet,isensee2021nnu}.

\subsection{Attention and Context Modeling}

As the field matured, the focus shifted from local volumetric features to richer context and more selective feature fusion. Multi-scale 3D CNNs with structured refinement improved lesion delineation in heterogeneous MRI volumes~\cite{kamnitsas2017multiscalecrf}. Attention U-Net introduced decoder-guided attention gates for skip connections~\cite{oktay2018attentionunet}, while channel and spatial attention modules such as SE and CBAM offered lightweight alternatives for feature recalibration~\cite{hu2018senet,woo2018cbam}. More recently, transformer-inspired models such as UNETR and Swin-UNet incorporated stronger long-range reasoning, but often at substantially higher computational cost~\cite{hatamizadeh2022unetr,cao2022swinunet}.

\subsection{Efficiency, Normalization, and Robustness}

Efficiency-oriented network design has become increasingly important in volumetric segmentation. Depthwise separable convolutions, popularized by MobileNet~\cite{howard2017mobilenets}, substantially reduce parameter count and computation, yet their systematic use in 3D medical segmentation remains limited. In parallel, normalization choices have been re-examined because Batch Normalization can be unstable under the small batch sizes typical of volumetric training~\cite{ioffe2015batchnorm}. Group Normalization~\cite{wu2018groupnorm} and related alternatives have therefore become common in 3D CNNs, and domain adaptation work has highlighted the importance of handling scanner and protocol variability explicitly~\cite{ganin2016domain,kamnitsas2017unsupervised,rawat2022layernorm}. BraTS-GoAT and related work further emphasize that robustness under acquisition variability remains an open challenge~\cite{satushe2025bratsgoat}.
\begin{table}[!htbp]
\centering
\caption{Summary of representative related work and the remaining gaps motivating DALight-3D}
\label{tab:lit_summary}
\begin{tabular}{|p{2.0cm}|p{2.6cm}|p{3.4cm}|p{3.4cm}|}
\hline
\textbf{Representative work} & \textbf{Problem addressed} & \textbf{Main idea / method} & \textbf{Remaining limitation} \\
\hline
Gordillo et al.~\cite{gordillo2013survey} & Delineating tumors from heterogeneous MRI with hand-crafted rules & Clustering, deformable models, intensity- and shape-based heuristics & Sensitive to noise, scanner variability, and large appearance changes; weak generalization \\
\hline

Havaei et al.~\cite{havaei2017brainseg} & Replace hand-crafted features with learned representations & Patch-based / fully convolutional CNN feature learning for tumor segmentation & Limited long-range volumetric context; often multi-stage and computationally expensive \\
\hline

Ronneberger et al.; Cicek et al.; Milletari et al.~\cite{ronneberger2015unet,cicek20163dunet,milletari2016vnet} & Dense biomedical and volumetric segmentation & Encoder--decoder architectures with skip connections and volumetric convolutions & Strong performance but rapidly increasing compute and memory cost in 3D \\
\hline

Wang et al.; Andermatt et al.~\cite{wang2019cascaded,andermatt2018multiplanar} & Improve volumetric context and receptive field & Cascaded 3D CNNs, anisotropic kernels, orthogonal-plane fusion & Better context but higher pipeline complexity and limited acquisition adaptation \\
\hline

Isensee et al.~\cite{isensee2018nonewnet,isensee2021nnu} & Strong performance without architectural novelty & Careful configuration, data processing, and training heuristics & Robust baselines, but not explicitly lightweight or conditioned for acquisition variability \\
\hline

Kamnitsas et al.; Oktay et al.; Hu et al.; Woo et al.~\cite{kamnitsas2017multiscalecrf,oktay2018attentionunet,hu2018senet,woo2018cbam} & Better feature selection and context modeling & Multi-scale CNNs, CRF refinement, attention-gated skips, channel/spatial recalibration & Improved context/fusion, but often with extra modules and limited efficiency emphasis \\
\hline

Hatamizadeh et al.; Cao et al.~\cite{hatamizadeh2022unetr,cao2022swinunet} & Capture longer-range dependencies & Self-attention and transformer-based encoder--decoder designs & Strong contextual modeling but high computational cost for 3D volumes \\
\hline

Ganin et al.; Kamnitsas et al.; Satushe et al.~\cite{ganin2016domain,kamnitsas2017unsupervised,satushe2025bratsgoat} & Robustness across scanners, sites, and tumor types & Domain-invariant learning or explicitly robustness-focused evaluation & Generalization is highlighted, but often not integrated into a compact segmentation backbone \\
\hline

Liu et al.; Lopez-Ramirez et al.~\cite{liu2022convnext,lopezramirez2026multiscaleconvnext} & Efficiency and improved robustness in modern backbones & ConvNeXt-style modernization and multi-scale feature extraction & Strong efficiency trend, but no single compact 3D design jointly unifies conditioning, low-cost long-range context, and adaptive skip fusion \\
\hline

Akter et al.~\cite{akter2024robust} & Improve practical MRI segmentation and classification robustness & CNN/U-Net style integration with clinically motivated evaluation & Better practical robustness, but limited explicit conditioned adaptation and parameter-efficiency focus \\
\hline

Rawat et al.~\cite{rawat2022layernorm} & Stabilize 3D segmentation training under small-batch MRI settings & Replace or compare normalization strategies in 3D U-Net segmentation & Improves optimization behavior, but does not by itself solve context modeling or efficiency \\
\hline

Henry et al.~\cite{henry2021selfensembled} & Push benchmark performance through stronger training schemes & Deep supervision, self-ensembling, and multi-model U-Net variants & High challenge performance, but greater training complexity and limited deployment efficiency \\
\hline
\end{tabular}
\end{table}

\subsection{Research Gap}

Taken together, the literature reveals three practical gaps that motivate DALight-3D: (i) lightweight 3D feature extraction, (ii) explicit feature conditioning under heterogeneous acquisition settings, and (iii) efficient long-range and skip-level context modeling without resorting to heavy transformer-style computation. DALight-3D addresses these gaps by integrating SepConv, identifier-conditioned normalization, CSA, and SSFB within a single compact encoder--decoder framework. Table~\ref{tab:lit_summary} condenses this progression and highlights the remaining limitations that directly motivate the proposed design.

\begin{figure}[t]
\centering
\includegraphics[width=\linewidth]{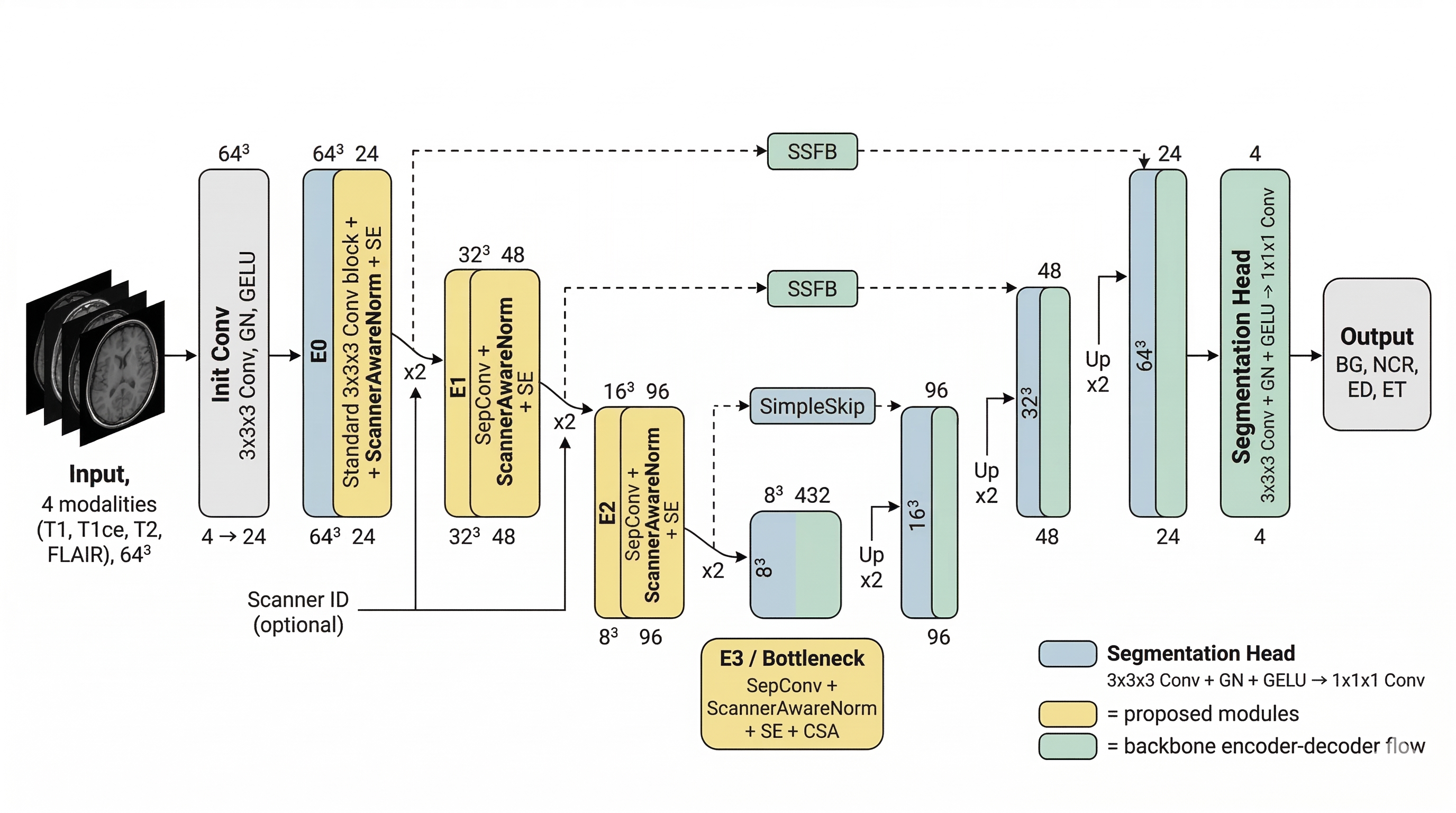}
\caption{DALight-3D Architecture: Standard convolution is used in the first encoder block, separable convolution in deeper blocks, CSA in E2 and E3, and SSFB in decoder skip fusion}
\label{fig:arch}
\end{figure}

\section{Proposed Methodology: DALight-3D}
\label{sec:method}

This section presents the proposed \textbf{DALight-3D} network for multi-modal brain tumor segmentation. The model is designed around two practical requirements that remain central in volumetric medical image analysis: reducing the computational cost of 3D segmentation and improving feature conditioning under heterogeneous MRI inputs. To this end, DALight-3D retains the strong encoder--decoder prior of 3D U-Net while introducing four targeted modifications: separable 3D convolution for efficient representation learning, conditioned normalization for feature calibration, cross-slice attention for low-cost contextual aggregation, and adaptive skip fusion for stronger decoder reconstruction.

\subsection{Problem Formulation}
\label{sec:problem}

Let
\begin{equation}
\mathbf{x} \in \mathbb{R}^{M \times D \times H \times W}
\end{equation}
denote a multi-modal MRI volume with $M=4$ modalities (T1, T1ce, T2, and FLAIR), and let
\begin{equation}
\mathbf{y} \in \{1,\dots,K\}^{D \times H \times W}, \qquad K=4,
\end{equation}
be the corresponding voxel-wise annotation over
\[
\mathcal{C}=\{\text{BG},\text{NCR},\text{ED},\text{ET}\},
\]
where BG, NCR, ED, and ET denote background, necrotic/non-enhancing tumor core, edema, and enhancing tumor, respectively. The segmentation task is to learn a mapping
\begin{equation}
f_{\theta}:(\mathbf{x},s) \mapsto \mathbf{p} \in [0,1]^{K \times D \times H \times W},
\end{equation}
where $\theta$ denotes the trainable parameters and $s \in \{0,\dots,S-1\}$ is an optional identifier. For each voxel $v \in \Omega$, the posterior probability of class $c$ is obtained from the predicted logits $\mathbf{z}$ as
\begin{equation}
p_c(v)=\frac{\exp(z_c(v))}{\sum_{c'=1}^{K}\exp(z_{c'}(v))}.
\end{equation}

DALight-3D is formulated as a four-stage encoder--decoder model,
\begin{equation}
\mathbf{f}^{(\ell)} = \mathrm{Enc}^{(\ell)}\big(\mathbf{f}^{(\ell-1)}; s\big), \qquad \ell=0,\dots,3,
\end{equation}
with $\mathbf{f}^{(-1)}$ denoting the initial projected input features, and
\begin{equation}
\mathbf{p}=\mathrm{Softmax}\!\left(\mathrm{Head}\!\left(\mathrm{Dec}\!\left(\mathbf{f}^{(0)},\mathbf{f}^{(1)},\mathbf{f}^{(2)},\mathbf{f}^{(3)}; s\right)\right)\right).
\end{equation}

\subsection{Preprocessing and Patch Sampling}
\label{sec:preprocess}

Following standard BraTS-style preprocessing, each modality is normalized independently and training is performed on fixed-size 3D patches. For each modality channel $m$, statistics are computed over non-background voxels and the input is normalized as
\begin{equation}
\tilde{x}_{m}(v) =
\begin{cases}
\dfrac{x_m(v) - \mu_m}{\sigma_m + \epsilon}, & x_m(v) > 0, \\
0, & \text{otherwise}.
\end{cases}
\end{equation}
Training patches are sampled with bias toward tumor voxels to increase the fraction of informative samples while keeping memory usage bounded. In the evaluated setting, the patch size is $64^3$.

\subsection{Architecture Overview}
\label{sec:arch-figure}

The design motivation follows directly from the research gaps identified in Section~\ref{sec:lit}: a compact 3D backbone can remain competitive if efficiency, conditioned normalization, context modeling, and skip fusion are addressed jointly rather than in isolation. The feature width follows
\begin{equation}
C_\ell = \min(C_0 \cdot 2^\ell, C_{\max}),
\end{equation}
where $C_0=24$ is the base width and $C_{\max}=384$ caps intermediate channel growth. In the evaluated model, the final encoder stage is widened to 432 channels to increase bottleneck capacity without widening the entire network.

As shown in Figure~\ref{fig:arch}, E0--E3 operate at resolutions $64^3$, $32^3$, $16^3$, and $8^3$, respectively. The first encoder stage uses full $3\times3\times3$ convolution to preserve expressive low-level features, whereas deeper stages use separable convolutions. Cross-slice attention is inserted only in E2 and E3, where the reduced spatial resolution makes attention more efficient. During decoding, transposed convolutions recover spatial resolution; the first skip path uses simple fusion for stability, whereas the remaining skip paths use SSFB.

\begin{table}[!htbp]
\centering
\caption{Feature Dimensions: Input $64^3$, base width $C_0=24$, and bottleneck width 432}
\label{tab:dims}
\begin{tabular}{|l|c|c|}
\hline
Stage & Resolution & Channels \\
\hline
Input & $64^3$ & 4 \\
\hline
After Init & $64^3$ & 24 \\
\hline
E0, E1, E2, E3 & $64^3$, $32^3$, $16^3$, $8^3$ & 24, 48, 96, 432 \\
\hline
Bottleneck & $8^3$ & 432 \\
\hline
Decoder (D0, D1, D2) & $8^3$, $16^3$, $32^3$ & 432, 96, 48 \\
\hline
After final decoder & $64^3$ & 24 \\
\hline
Output & $64^3$ & 4 \\
\hline
\end{tabular}
\end{table}

\subsection{Encoder and Lightweight Feature Extraction}
\label{sec:encoder}

The encoder begins with an initial projection layer
\begin{equation}
\tilde{\mathbf{f}}^{(0)} = \mathrm{GELU}\big( \mathrm{GN}( \mathrm{Conv}_{3\times3\times3}(\mathbf{x}) ) \big),
\end{equation}
which maps the input modalities to $C_0$ channels. The first encoder feature map is then obtained as
\begin{equation}
\mathbf{f}^{(0)} = \mathrm{Block}^{(0)}(\tilde{\mathbf{f}}^{(0)}),
\end{equation}
where $\mathrm{Block}^{(0)}$ uses full $3\times3\times3$ convolution. For deeper levels,
\begin{equation}
\mathbf{f}^{(\ell)} = \mathrm{Block}^{(\ell)}\big( \mathrm{Downsample}(\mathbf{f}^{(\ell-1)}) \big), \qquad \ell = 1,2,3,
\end{equation}
where downsampling is implemented by a strided $3\times3\times3$ convolution with stride $2$.

\subsection{LightweightBlock}
\label{sec:lightweightblock}

The \textbf{LightweightBlock} is the main residual unit of DALight-3D. Given an input tensor $\mathbf{h} \in \mathbb{R}^{C_{\mathrm{in}} \times D \times H \times W}$, the block applies two convolution--normalization transformations followed by channel recalibration:
\begin{equation}
\mathbf{z} = \mathrm{SE}\!\Big( \mathrm{Norm}_2\big( \mathrm{Conv}_2( \phi( \mathrm{Norm}_1( \mathrm{Conv}_1(\mathbf{h}) ) ) ) \big) \Big),
\end{equation}
where $\phi$ denotes GELU~\cite{hendrycks2016gelu}, $\mathrm{Norm}_1$ and $\mathrm{Norm}_2$ denote ScannerAwareNorm (or GroupNorm when conditioning is unavailable), and $\mathrm{Conv}_i$ is standard convolution in E0 and SepConv in E1--E3. Channel recalibration is implemented with an SE-style block~\cite{hu2018senet}. For encoder stages E2 and E3, contextual refinement is added by
\begin{equation}
\mathbf{z} \leftarrow \mathbf{z} + \mathrm{CrossSliceAttn}(\mathbf{z}).
\end{equation}
A residual projection then produces the block output:
\begin{equation}
\mathbf{o} = \phi\big(\mathbf{z} + \mathrm{Proj}(\mathbf{h})\big),
\end{equation}
where $\mathrm{Proj}$ is identity when shapes match and a $1\times1\times1$ projection otherwise.

\subsection{Separable 3D Convolution (SepConv)}
\label{sec:sepconv}

A standard 3D convolution requires $\mathcal{O}(C_{\mathrm{in}} C_{\mathrm{out}} k^3)$ parameters, which rapidly increases memory and computation as the feature width grows. DALight-3D therefore uses depthwise separable 3D convolution~\cite{howard2017mobilenets} in the deeper encoder and decoder blocks. Given an input feature map $\mathbf{u}$, SepConv is written as
\begin{equation}
\mathbf{v} = \mathcal{W}_{\mathrm{dw}} \ast \mathbf{u},
\end{equation}
\begin{equation}
\mathbf{y} = \mathcal{W}_{\mathrm{pw}} \ast \mathbf{v},
\end{equation}
where $\mathcal{W}_{\mathrm{dw}}$ performs channel-wise spatial filtering and $\mathcal{W}_{\mathrm{pw}}$ performs $1\times1\times1$ channel mixing. The corresponding parameter cost becomes $C_{\mathrm{in}}k^3 + C_{\mathrm{in}}C_{\mathrm{out}}$.

\subsection{Scanner-Aware Normalization (ScannerAwareNorm)}
\label{sec:scannorm}

A recurring challenge in multi-site MRI is that feature statistics can shift across scanners and acquisition protocols. DALight-3D introduces \textbf{ScannerAwareNorm}, a scanner-conditioned extension of Group Normalization~\cite{wu2018groupnorm}. For a feature tensor $\mathbf{x}$, we first compute
\begin{equation}
\hat{\mathbf{x}} = \mathrm{GN}(\mathbf{x}),
\end{equation}
then modulate the normalized activations using conditioned affine parameters:
\begin{equation}
\mathbf{y} = \gamma_s \odot \hat{\mathbf{x}} + \beta_s,
\end{equation}
where $\gamma_s = \mathbf{E}_{\gamma}(s)$ and $\beta_s = \mathbf{E}_{\beta}(s)$ are learnable embeddings indexed by identifier $s$. When identifier information is unavailable, the module falls back to learned default affine parameters.

\subsection{Cross-Slice Attention (CSA)}
\label{sec:csa}

Although 3D convolution captures local neighborhoods well, it does not explicitly model long-range dependencies across distant slices. DALight-3D therefore employs \textbf{Cross-Slice Attention (CSA)} in the deeper encoder stages. Given
\begin{equation}
\mathbf{x} \in \mathbb{R}^{B \times C \times D \times H \times W},
\end{equation}
we first pool over the in-plane dimensions to obtain
\begin{equation}
\mathbf{p} = \mathrm{Pool}_{H,W}(\mathbf{x}) \in \mathbb{R}^{B \times C \times D}.
\end{equation}
Low-rank query, key, and value projections are then computed as
\begin{equation}
Q = W_q \mathbf{p}, \qquad K = W_k \mathbf{p}, \qquad V = W_v \mathbf{p}.
\end{equation}
The slice-attention matrix is
\begin{equation}
A = \mathrm{softmax}\left( \frac{Q^\top K}{\sqrt{d}} \right),
\end{equation}
and the refined output is given by
\begin{equation}
\mathrm{CSA}(\mathbf{x}) = \mathbf{x} + \mathrm{Broadcast}(W_o V A).
\end{equation}
Because attention is computed along the slice axis rather than across all voxels, the effective complexity is reduced from $O((DHW)^2)$ to $O(D^2)$.

\subsection{Decoder and Multi-scale Feature Fusion}
\label{sec:decoder}

The decoder reconstructs the segmentation map progressively through transposed-convolution upsampling. At each resolution level, the upsampled decoder representation is fused with its corresponding encoder feature map and then refined by a LightweightBlock without CSA. The first skip path uses simple fusion, whereas the remaining skip paths use the proposed SSFB module. When exact spatial alignment is not available, trilinear interpolation is used before fusion.

\subsection{Skip and Spatial Feature Blend (SSFB)}
\label{sec:ssfb}

The \textbf{Skip and Spatial Feature Blend (SSFB)} module improves skip fusion beyond naive concatenation. Let $\mathbf{f}_{\mathrm{dec}}$ and $\mathbf{f}_{\mathrm{enc}}$ denote the aligned decoder and encoder feature maps. SSFB combines two complementary pathways.

\paragraph{Low-rank attention pathway.}
A low-rank cross-attention operation computes
\begin{equation}
\mathbf{o}_{\mathrm{attn}} = W_{\mathrm{out}}(VA),
\end{equation}
where $A$ is formed from low-rank query and key projections of the decoder and encoder features.

\paragraph{Channel-gating pathway.}
A channel-reweighting branch computes
\[
\mathbf{o}_{\mathrm{gate}} = \mathbf{f}_{\mathrm{enc}} \odot \sigma\big( \mathrm{MLP}( [\mathrm{GAP}(\mathbf{f}_{\mathrm{dec}}); \mathrm{GAP}(\mathbf{f}_{\mathrm{enc}})] ) \big),
\]
where $\mathrm{GAP}$ denotes global average pooling and $\sigma$ is the sigmoid nonlinearity.

The two branches are blended through a learnable scalar:
\begin{equation}
\mathbf{m} = \alpha \, \mathbf{o}_{\mathrm{attn}} + (1-\alpha) \, \mathbf{o}_{\mathrm{gate}},
\end{equation}
where $\alpha \in (0,1)$ is optimized during training. The final SSFB output is
\begin{equation}
\mathrm{SSFB}(\mathbf{f}_{\mathrm{dec}}, \mathbf{f}_{\mathrm{enc}}) = \phi\!\left(\mathrm{GN}\!\left(\mathrm{Conv}_{3\times3\times3}([\mathbf{f}_{\mathrm{dec}};\mathbf{m}])\right)\right).
\end{equation}

\subsection{Segmentation Head}
\label{sec:head}

The final decoder representation is converted to voxel-wise logits through a lightweight segmentation head:
\begin{equation}
\mathbf{z} = \mathrm{Conv}_{1\times1\times1}\big(\phi(\mathrm{GN}(\mathrm{Conv}_{3\times3\times3}(\mathbf{f}_{\mathrm{dec}})))\big).
\end{equation}
A softmax layer then produces the class-probability volume $\mathbf{p}$.

\subsection{Illustrative Prediction Visualization}
\label{sec:illustrative-prediction}

\begin{figure}[t]
\centering
\includegraphics[width=\linewidth]{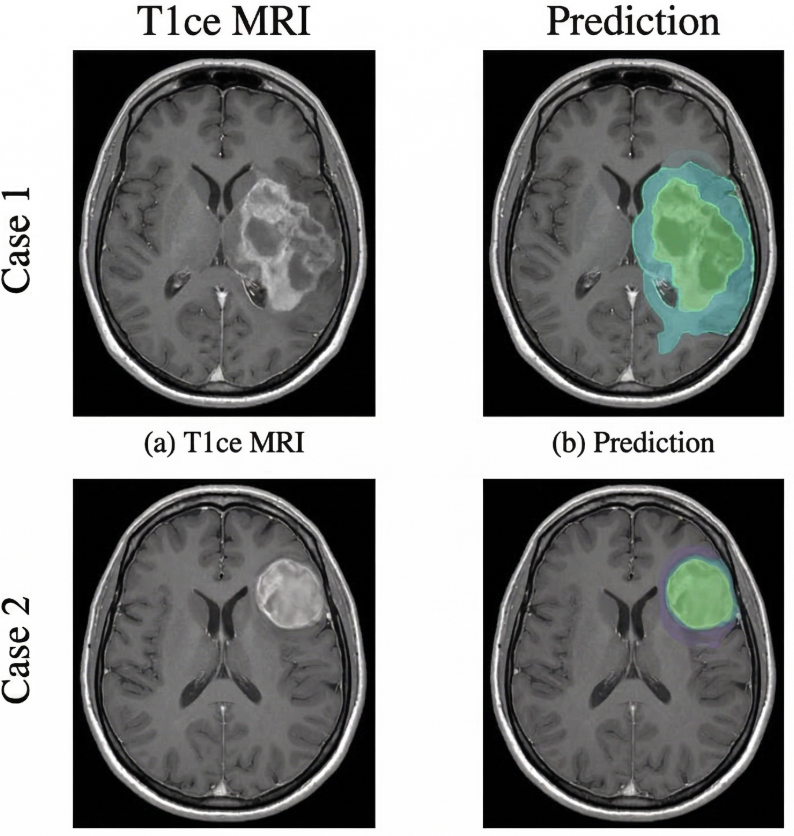}
\caption{Illustrative Prediction: Representative MRI slices and tumor prediction overlays targeted by DALight-3D}
\label{fig:illustrative-prediction}
\end{figure}

To complement the quantitative analysis, Figure~\ref{fig:illustrative-prediction} shows representative MRI slices together with illustrative tumor prediction overlays. Such visualizations help clarify the target output space of DALight-3D and provide an intuitive view of the lesion localization behavior expected from the model before quantitative evaluation is discussed.

\subsection{Training Objective}
\label{sec:loss}

The training objective combines overlap-sensitive Dice optimization with voxel-wise probabilistic supervision:
\begin{equation}
\mathcal{L} = \lambda_{\mathrm{Dice}} \mathcal{L}_{\mathrm{Dice}}(\mathbf{p}, \mathbf{y}) + \lambda_{\mathrm{CE}} \mathcal{L}_{\mathrm{CE}}(\mathbf{p}, \mathbf{y}),
\end{equation}
with
\begin{equation}
\lambda_{\mathrm{Dice}}=1, \qquad \lambda_{\mathrm{CE}}=0.5.
\end{equation}
Let $\mathcal{T}=\{\text{NCR},\text{ED},\text{ET}\}$ denote the tumor classes. The Dice term is defined as
\begin{equation}
\mathcal{L}_{\mathrm{Dice}} = 1 - \frac{1}{|\mathcal{T}|} \sum_{c \in \mathcal{T}} \frac{2\sum_{v \in \Omega} p_c(v) y_c(v) + \epsilon}{\sum_{v \in \Omega} p_c(v) + \sum_{v \in \Omega} y_c(v) + \epsilon},
\end{equation}
and the cross-entropy term is
\begin{equation}
\mathcal{L}_{\mathrm{CE}} = -\frac{1}{|\Omega|} \sum_{v \in \Omega} \sum_{c=1}^{K} y_c(v) \log p_c(v).
\end{equation}

\section{Evaluation}
\label{sec:eval}

\subsection{Experimental Setup}
\label{sec:experimental_setup}

Experiments were prepared locally and executed in Google Colab using an NVIDIA A100 GPU. Model development, training, evaluation, and figure generation were performed in a Python/PyTorch environment, and all reported model comparisons used the same runtime configuration so that the measured differences primarily reflect architecture rather than hardware. Unless otherwise stated, the results in this section correspond to Medical Segmentation Decathlon Task01\_BrainTumour under a 50-epoch training schedule.

\subsection{Dataset and Preprocessing}
\label{sec:dataset}

The dataset is Medical Segmentation Decathlon Task01\_BrainTumour~\cite{simpson2019msd}, a BraTS-derived multi-modal MRI benchmark containing T1, T1ce, T2, and FLAIR volumes with voxel-wise annotations for background (BG), necrotic/non-enhancing tumor core (NCR), peritumoral edema (ED), and enhancing tumor (ET). We follow the official training/validation partition. Each case is normalized independently using z-score normalization, and training is performed on tumor-centered $64^3$ patches with random flips and light intensity perturbations. The local dataset files used here do not expose harmonized scanner metadata, so ScannerAwareNorm is conditioned on eight deterministic proxy buckets defined as $\mathrm{hash}(\mathrm{case\_id}) \bmod 8$. Eight buckets were used as a lightweight fixed conditioning cardinality in the current implementation. Accordingly, the present experiments should be interpreted as evaluating identifier-conditioned normalization within this benchmark, not as a direct cross-scanner generalization study.

\subsection{Implementation and Training Protocol}
\label{sec:implementation}

All models are implemented in PyTorch and trained under matched settings to ensure a controlled comparison. We use AdamW~\cite{loshchilov2019adamw} with learning rate $5\times10^{-5}$, a cosine annealing scheduler~\cite{loshchilov2017sgdr}, and the hybrid Dice plus cross-entropy objective with $\lambda_{\mathrm{Dice}}=1$ and $\lambda_{\mathrm{CE}}=0.5$. Training uses batch size 1, validation every 2 epochs, and disabled mixed precision for numerical stability. The proposed DALight-3D uses base width $C_0=24$, bottleneck width 432, and SSFB rank 8, resulting in approximately 2.22M trainable parameters. The comparison models---standard 3D U-Net~\cite{cicek20163dunet}, Attention U-Net~\cite{oktay2018attentionunet}, Residual 3D U-Net~\cite{he2016resnet}, and V-Net~\cite{milletari2016vnet}---share the same training protocol and base width so that the observed differences primarily reflect architectural design. The ablation models were also run in the same uniform manner. 

\subsection{Component-Wise Ablation Protocol}
\label{sec:ablation_protocol}

To isolate the contribution of each proposed module, we define a component-wise ablation protocol in which one design element is removed at a time while all other training and architectural settings are kept fixed. This design separates the effect of efficiency-oriented operators from that of conditioned normalization, contextual aggregation, and adaptive skip fusion.

\begin{table}[!htbp]
\centering
\caption{Ablation Protocol: Each variant removes one proposed component while preserving the remaining architecture and training setup}
\label{tab:ablation_protocol}
\begin{tabular}{|l|l|}
\hline
Variant & Modification relative to full DALight-3D \\
\hline
Full model & All proposed components enabled \\
\hline
w/o SepConv & deeper separable convolutions \\
\hline
w/o ScannerAwareNorm & scanner-conditioned normalization \\
\hline
w/o CSA & cross-slice attention\\
\hline
w/o SSFB & SSFB \\
\hline
\end{tabular}
\end{table}

\subsection{Results}
\label{sec:results}

\subsubsection{Comparison with Baselines}

Table~\ref{tab:comparison} presents the primary quantitative comparison across all evaluated models. This experiment measures whether the proposed architecture improves the accuracy--efficiency operating point relative to established CNN baselines. The table reports mean Dice as the primary segmentation metric, total trainable parameters as a measure of model complexity, and Dice per million parameters as a compact indicator of parameter efficiency. Within this benchmark, DALight-3D attains the highest mean Dice (0.727) while also being the most compact model in the study (2.22M parameters). Relative to Residual 3D U-Net, DALight-3D improves mean Dice by 0.017 while reducing parameters by approximately 30.7\%. It also exceeds Standard 3D U-Net, Attention U-Net, and V-Net by 0.040, 0.043, and 0.022 Dice, respectively. The Dice-per-million-parameters values further support the efficiency of the proposed design.

\begin{table}[!htbp]
\centering
\caption{Baseline Comparison with Mean Dice and Parameter count}
\label{tab:comparison}
\begin{tabular}{|l|c|c|c|}
\hline
Method & Mean Dice $\uparrow$ & Params (M) & Dice/M \\
\hline
Standard 3D U-Net & 0.687 & 3.15 & 0.22 \\
\hline
Attention U-Net & 0.683 & 3.17 & 0.22 \\
\hline
Residual 3D U-Net & 0.710 & 3.20 & 0.22 \\
\hline
V-Net & 0.704 & 4.99 & 0.14 \\
\hline
\textbf{DALight-3D} & \textbf{0.727} & \textbf{2.22} & \emph{0.33} \\
\hline
\end{tabular}
\end{table}

\begin{figure}[t]
\centering
\includegraphics[width=\linewidth]{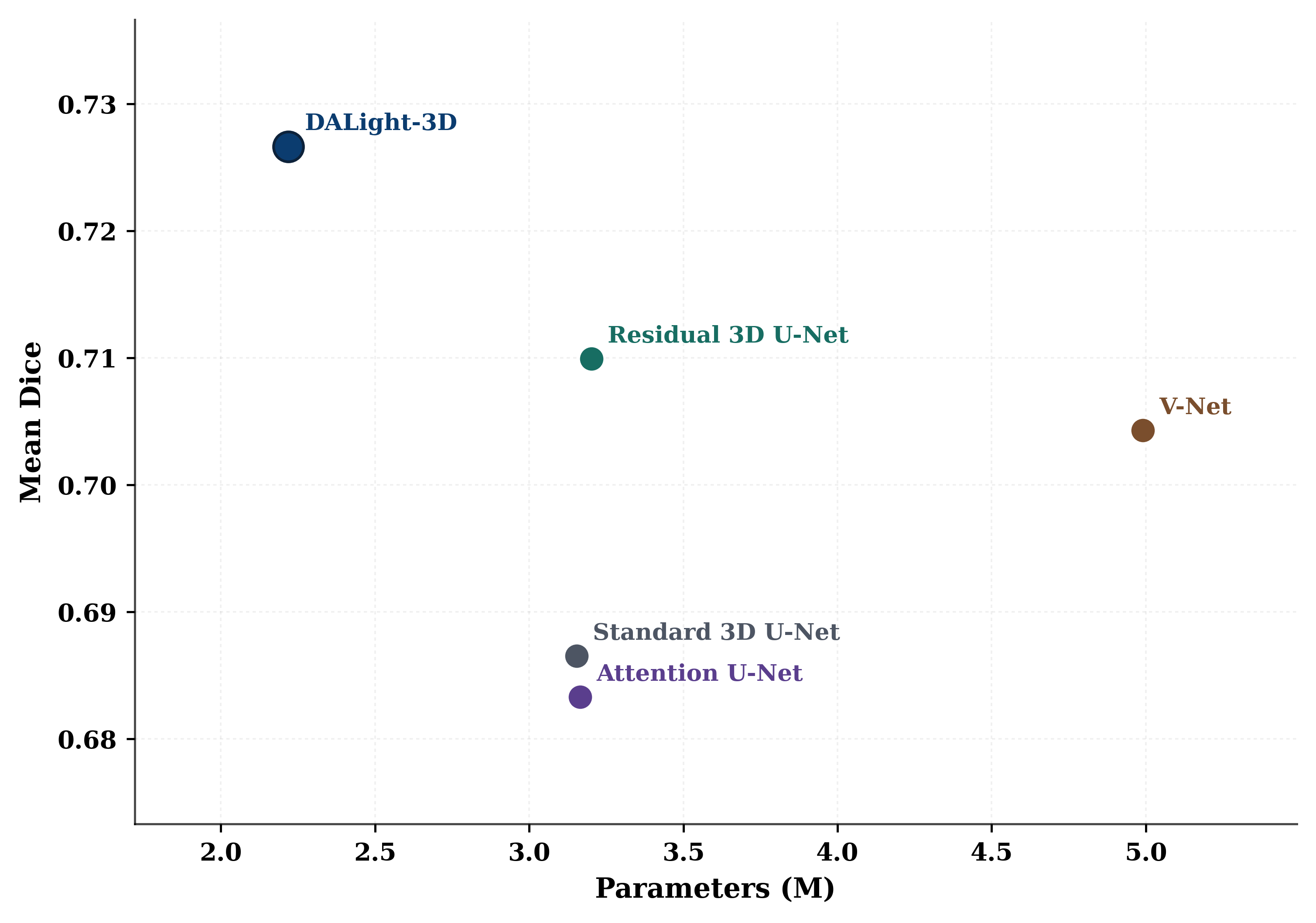}
\caption{Dice-Parameter Trade-Off: Mean Dice versus parameter count (millions)}
\label{fig:dice_params}
\end{figure}

\begin{figure}[t]
\centering
\includegraphics[width=\linewidth]{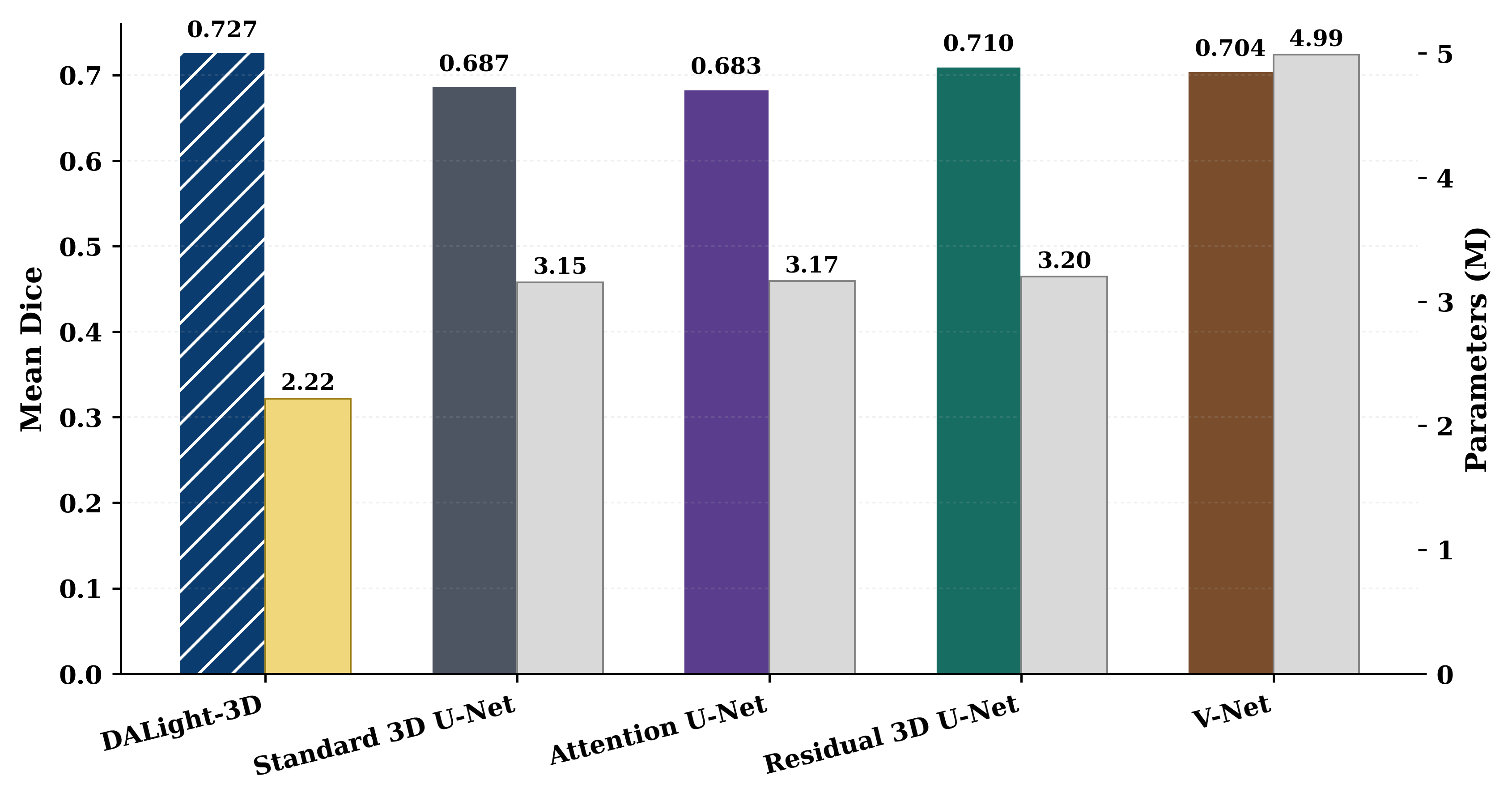}
\caption{Dice and Parameter Comparison: Side-by-side comparison across DALight-3D and the baseline models}
\label{fig:compare_dice_params_bar}
\end{figure}

Figure~\ref{fig:dice_params} and Figure~\ref{fig:compare_dice_params_bar} visualize the same relationship from two complementary perspectives. The scatter plot measures the global accuracy--efficiency trade-off by mapping model size against segmentation quality, where the upper-left region corresponds to high Dice at low parameter count. DALight-3D occupies this region, achieving the highest Dice while remaining substantially smaller than the larger baseline models. The bar chart presents the same result in a more direct pairwise comparison.

\subsubsection{Learning Curves}

\begin{figure}[t]
\centering
\includegraphics[width=\linewidth]{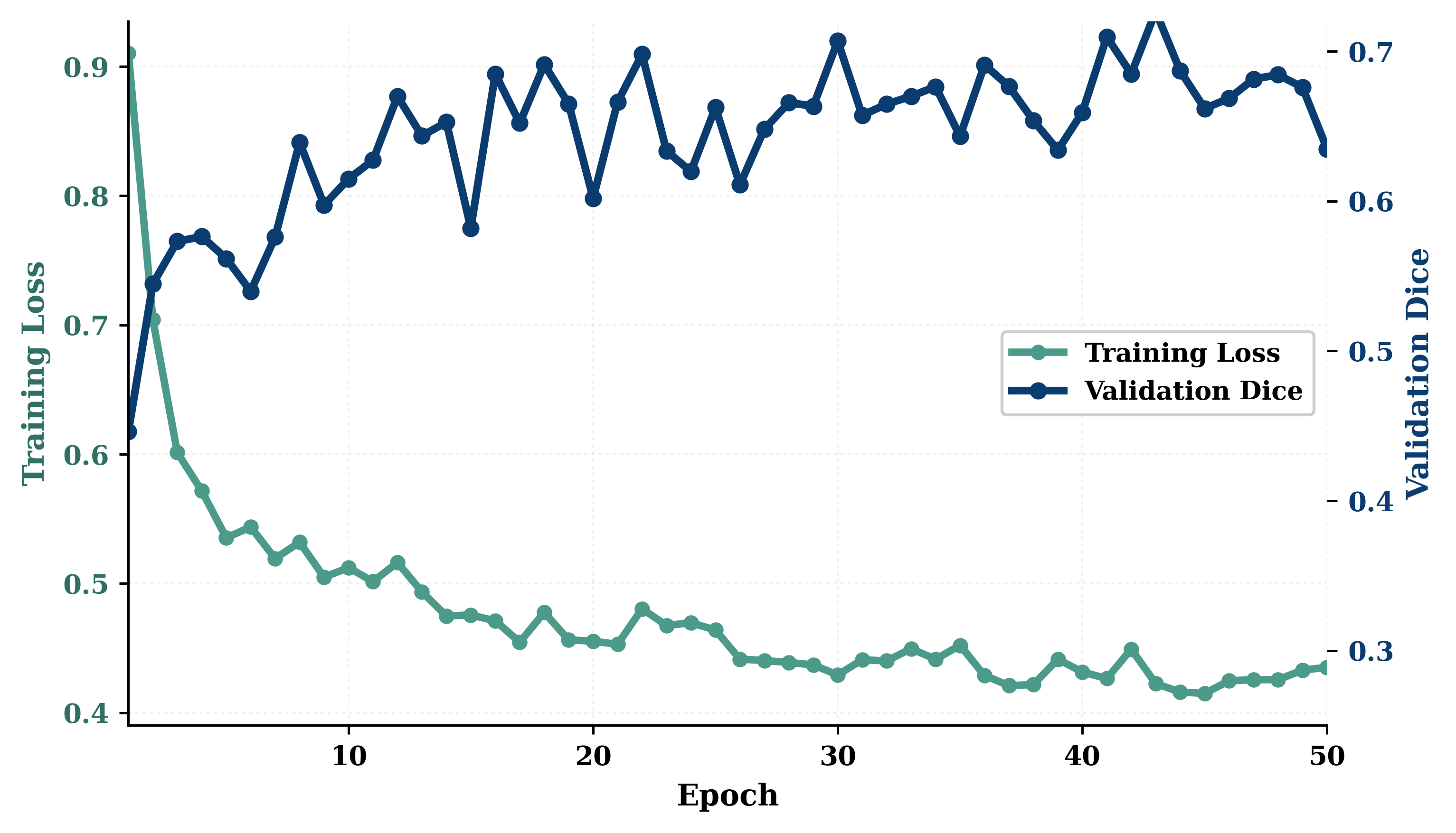}
\caption{DALight-3D Learning Curve: Training loss and validation Dice across the training run}
\label{fig:curves}
\end{figure}

The learning-curve analysis assesses whether the observed gains arise from stable optimization rather than isolated endpoint fluctuations. Figure~\ref{fig:curves} jointly reports DALight-3D training loss and validation Dice over the training schedule. This experiment measures whether the proposed model trains smoothly and continues to improve throughout optimization. The loss decreases consistently while validation Dice improves across the run, indicating stable optimization and continued extraction of useful structure over the observed epoch range. The absence of late-stage collapse or abrupt divergence suggests that the proposed lightweight design does not compromise trainability.

\begin{figure}[t]
\centering
\includegraphics[width=\linewidth]{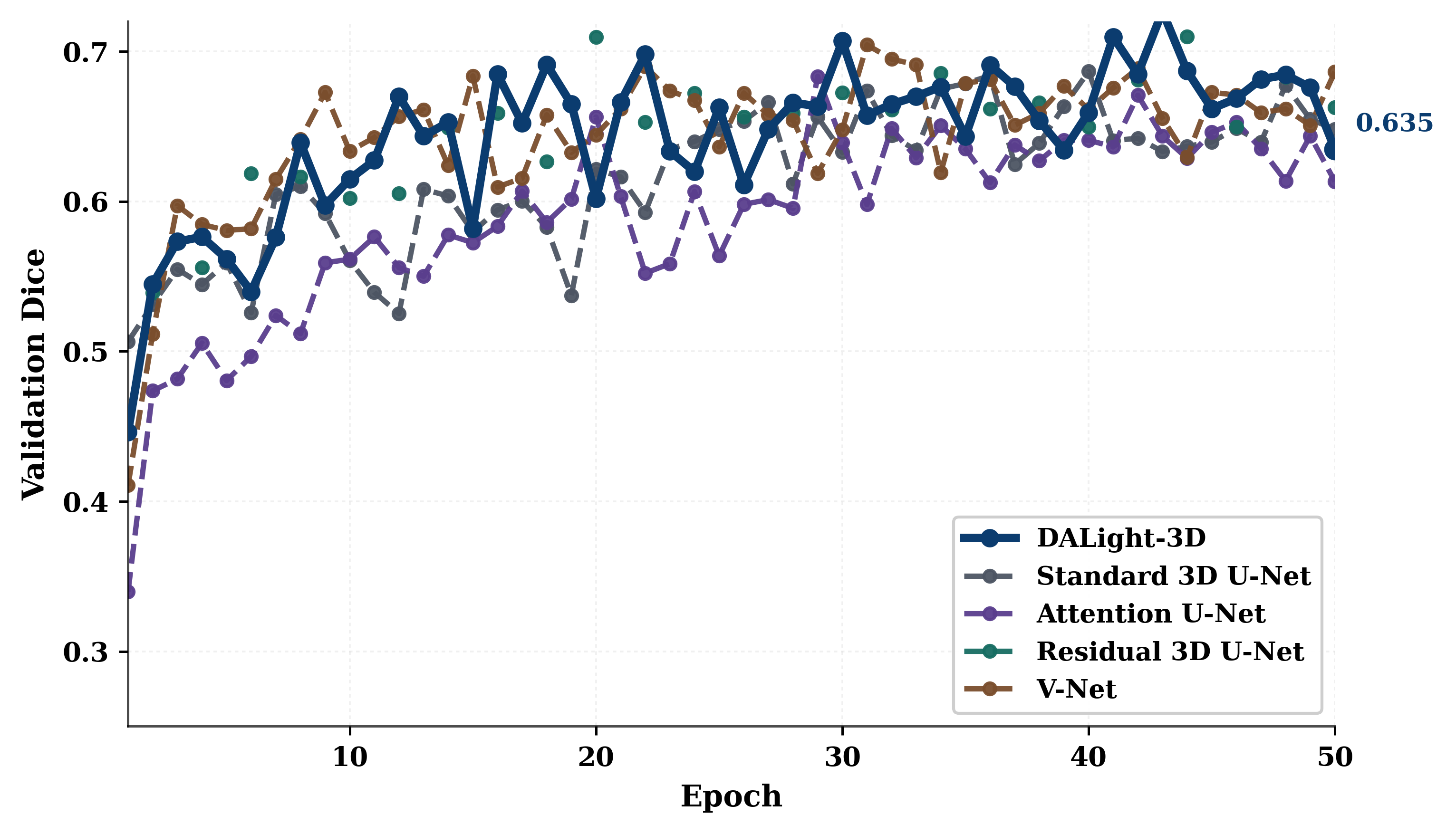}
\caption{Validation Dice Comparison: Validation Dice trajectories for DALight-3D and the baseline models}
\label{fig:compare_val_dice}
\end{figure}

\begin{figure}[t]
\centering
\includegraphics[width=\linewidth]{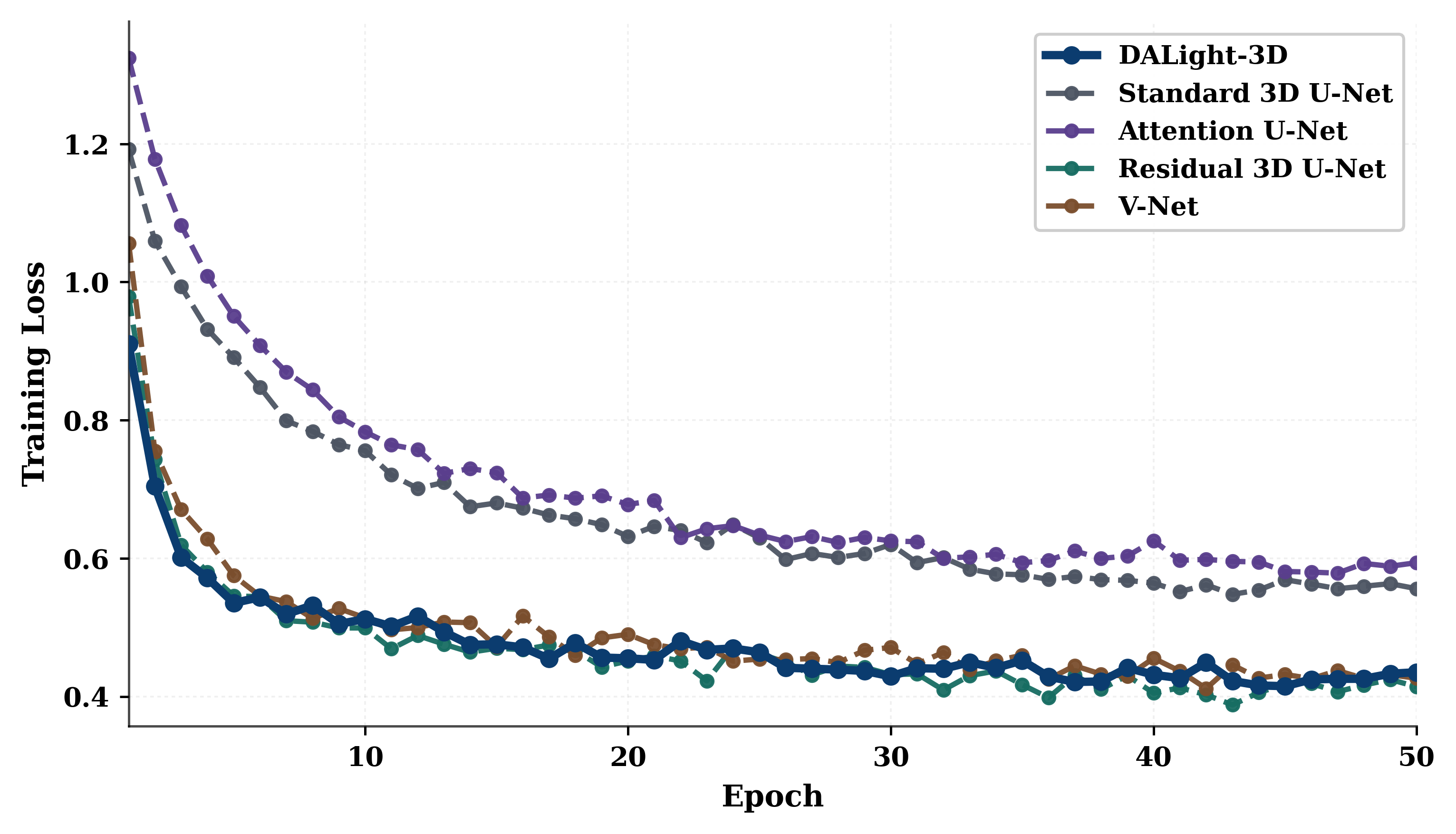}
\caption{Training Loss Comparison: Training loss trajectories for DALight-3D and the baseline models}
\label{fig:compare_train_loss}
\end{figure}

Figure~\ref{fig:compare_val_dice} compares validation Dice trajectories across DALight-3D and the baseline models. This experiment measures how rapidly and consistently each model converts training progress into validation improvement. DALight-3D remains competitive throughout training and finishes with the highest validation Dice. Figure~\ref{fig:compare_train_loss} complements this result by showing that the proposed method achieves smooth and stable optimization despite its lower parameter budget. Together, these curves indicate that the observed performance-efficiency trade-off is not an artifact of unstable training.

\begin{table}[!htbp]
\centering
\caption{Ablation Results: Component-wise results for DALight-3D}
\label{tab:ablation_results}
\begin{tabular}{|l|c|c|c|c|}
\hline
Variant & Epochs & Mean Dice $\uparrow$ & Params (M)\\
\hline
Full DALight-3D & 50 & 0.727 & 2.22\\
\hline
w/o SepConv & 50 & 0.667 & 9.20\\
\hline
w/o ScannerAwareNorm & 50 & 0.678 & 2.19\\
\hline
w/o CSA & 50 & 0.670 & 1.92 \\
\hline
w/o SSFB & 50 & 0.699 & 2.21 \\
\hline
\end{tabular}
\end{table}

\begin{figure}[t]
\centering
\includegraphics[width=\linewidth]{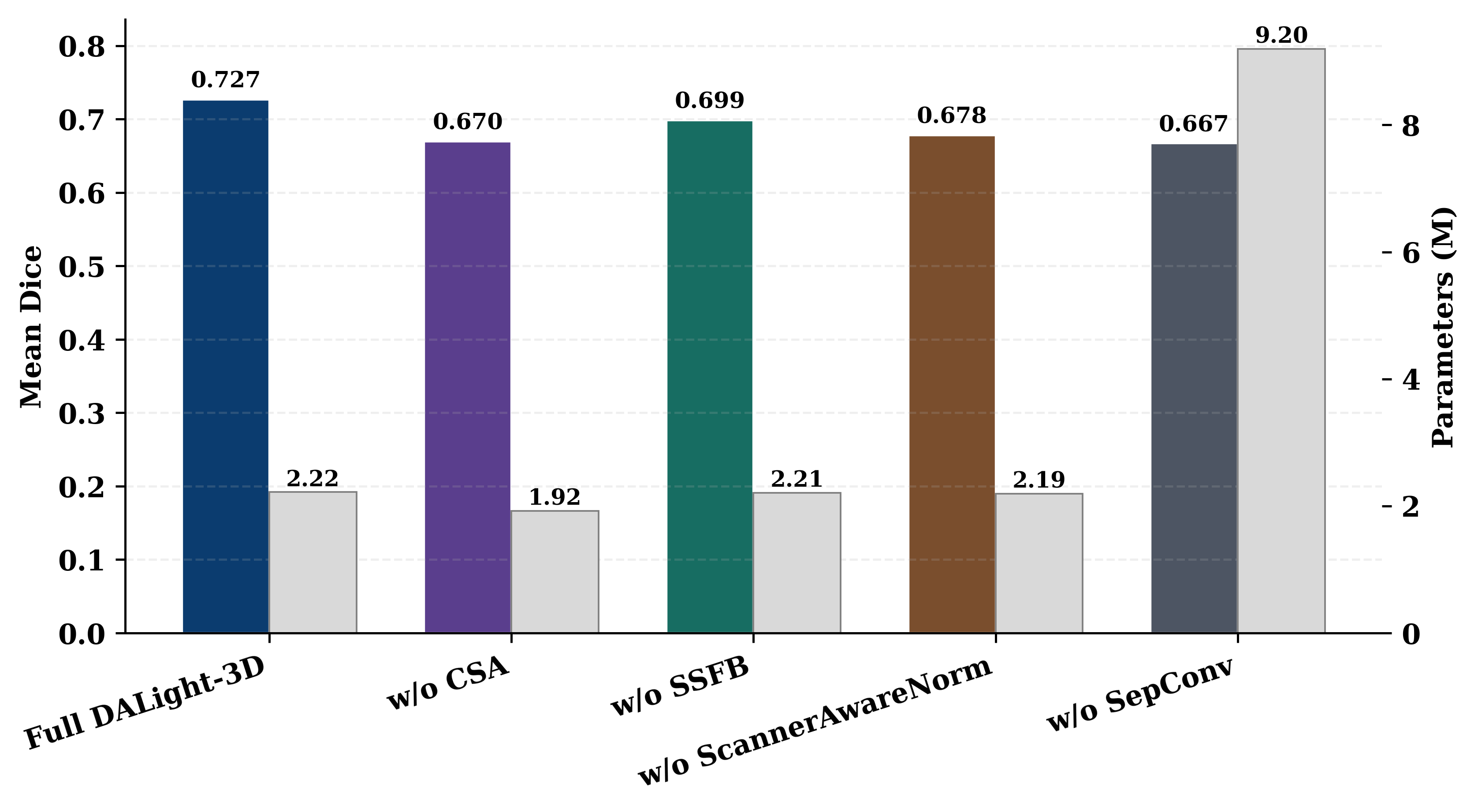}
\caption{Ablation Trade-Off: Effect of removing each component on Dice and parameter count}
\label{fig:ablation_dice_params}
\end{figure}

\subsubsection{Component-Wise Ablation Analysis}

Table~\ref{tab:ablation_results} and Figure~\ref{fig:ablation_dice_params} summarize the completed ablation runs. This experiment measures the contribution of each proposed component by removing one module at a time while keeping the remaining setup fixed. The full DALight-3D reference and the ablation variants each correspond to the same total training duration, although only the final 25 epochs were retained in the saved ablation histories. Every component removal lowers Dice relative to the full model, which indicates that the final gain is not dominated by a single isolated modification.

The largest degradation is observed when SepConv is removed ($-0.060$ Dice), and this variant also expands the parameter count from 2.22M to 9.20M. This result suggests that SepConv is not merely a parameter-reduction device, but a structural choice that allows the network to allocate capacity more effectively across stages. By replacing dense volumetric convolution with depthwise spatial filtering followed by pointwise channel mixing, the model preserves local receptive-field modeling while avoiding the aggressive parameter growth that can make lightweight 3D backbones harder to optimize under limited training budgets. In this sense, the gain associated with SepConv appears to arise from both efficiency and regularization: the model remains compact enough to train stably while still retaining sufficient expressivity for volumetric tumor delineation.

Removing CSA also causes a pronounced drop ($-0.057$ Dice), which supports the role of low-cost inter-slice context modeling. This finding is consistent with the observation that tumor extent and morphology are distributed across multiple adjacent slices, especially for diffuse regions and irregular boundaries. Standard 3D convolution captures local neighborhoods, but it does not explicitly model longer-range slice-to-slice interactions once the feature hierarchy has been compressed. CSA appears to compensate for this limitation by allowing deeper encoder features to exchange information along the depth axis at reduced spatial resolution, where the cost of attention is manageable. The ablation therefore indicates that a meaningful part of the performance gain comes from better contextual aggregation rather than from channel capacity alone.

Replacing ScannerAwareNorm with GroupNorm leads to a further $-0.049$ Dice, suggesting that conditioned feature modulation remains beneficial in this setup. Because the present implementation uses deterministic proxy identifiers rather than curated scanner metadata, this result should be interpreted cautiously; however, it still indicates that lightweight conditioning of feature statistics is useful when the input distribution is heterogeneous. A plausible explanation is that the conditioned affine parameters help the network adapt intermediate activations to recurring case-level intensity patterns that are not fully normalized away by preprocessing. The gain is therefore consistent with the broader hypothesis that normalization can serve not only as an optimization aid, but also as a mechanism for modest distribution-aware calibration inside a compact 3D backbone.

The smallest but still consistent drop is observed for SSFB ($-0.028$ Dice), indicating that adaptive skip fusion improves reconstruction quality on top of the other modules. This more moderate reduction is also informative: once feature extraction, normalization, and deep context modeling are already strengthened, the marginal role of skip fusion is naturally narrower. Even so, the positive contribution of SSFB suggests that decoder reconstruction benefits from more selective integration of encoder detail than naive skip concatenation provides. In particular, the combination of low-rank attention and channel gating appears to help the decoder reuse high-resolution information without indiscriminately passing all encoder activations forward. Overall, the ablation study supports the view that DALight-3D benefits from the interaction of all four components, with the strongest effects arising from efficiency-aware convolution and explicit deep contextual modeling.

\subsubsection{Per-Class Performance and Confusion Matrix}

\begin{figure}[t]
\centering
\includegraphics[width=\linewidth]{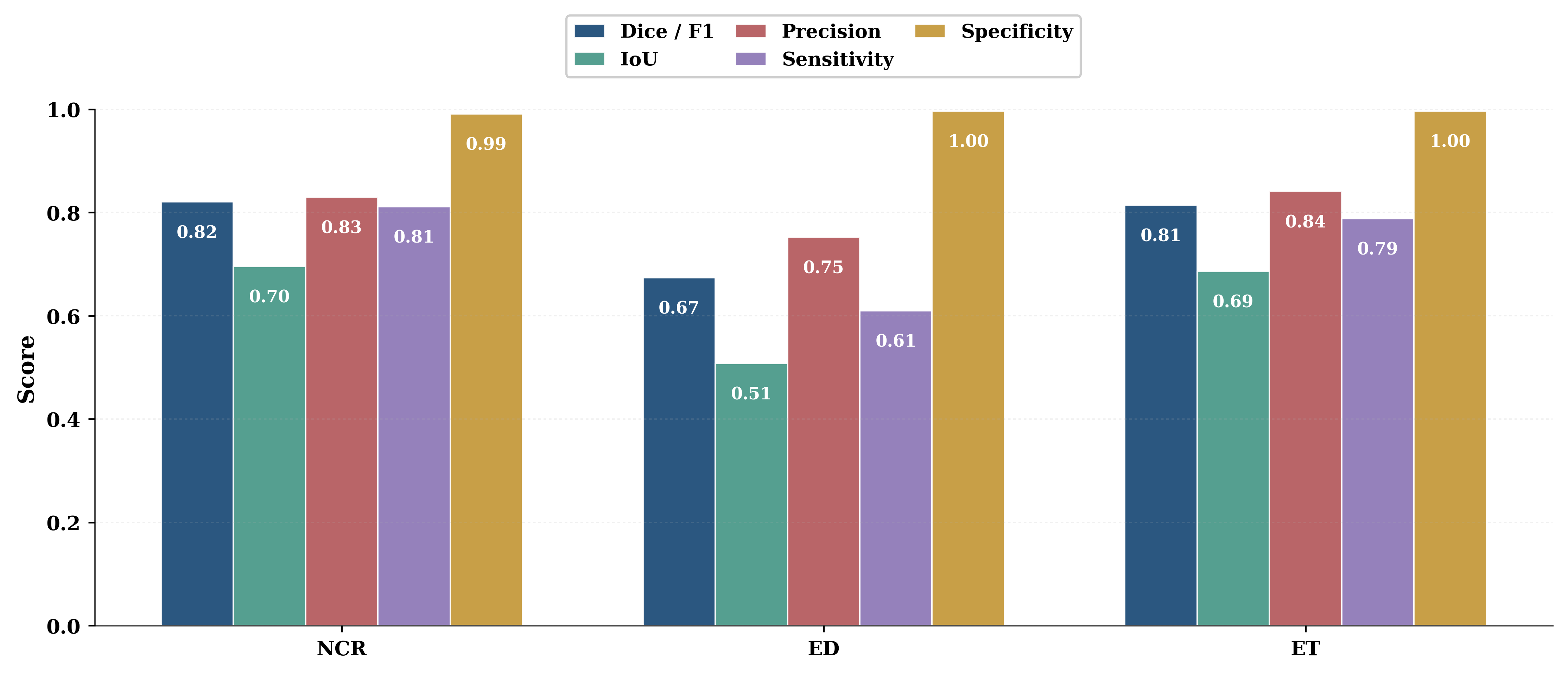}
\caption{Per-Class Performance: Dice/F1, IoU, precision, sensitivity, and specificity for NCR, ED, and ET}
\label{fig:perclass_full}
\end{figure}

The aggregate Dice score does not reveal which anatomical subregions are easier or harder for the model, so Figure~\ref{fig:perclass_full} breaks performance down by tumor class. This experiment measures whether the model behaves uniformly across tumor subregions or shows class-specific weaknesses. The per-class analysis reports voxel-level one-vs-rest metrics derived from the full validation-set confusion matrix: Dice/F1, IoU, precision, sensitivity, and specificity for NCR, ED, and ET, together with a tumor-class macro average. Here, NCR denotes necrotic and non-enhancing tumor core, ED denotes peritumoral edema, and ET denotes enhancing tumor. In this setting, Dice/F1 measures overlap between predicted and reference tumor voxels, IoU measures intersection relative to the union of prediction and ground truth, precision quantifies how many predicted positive voxels are correct, sensitivity quantifies how many true positive voxels are recovered, and specificity measures how effectively non-target voxels are rejected. NCR and ET achieve the strongest overall overlap quality, with Dice/F1 scores of 0.820 and 0.814, respectively. By contrast, ED remains the most challenging category, with Dice/F1 0.673 and IoU 0.508. This gap is clinically plausible because edema often exhibits diffuse boundaries and lower structural specificity than compact enhancing regions.

\begin{figure}[!t]
\centering
\includegraphics[width=\linewidth]{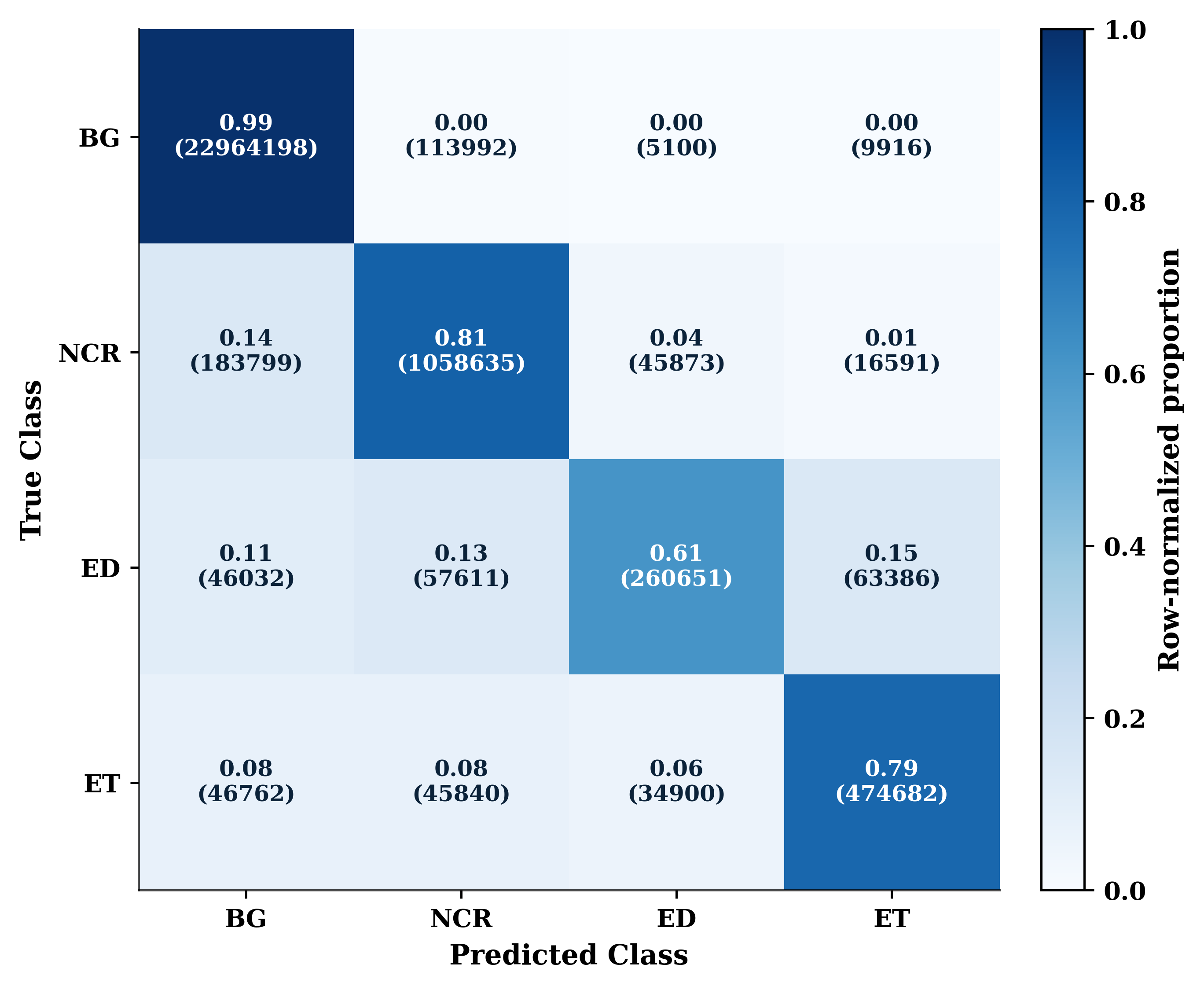}
\caption{Confusion Matrix: Voxel-wise confusion matrix for DALight-3D on the validation set}
\label{fig:confusion}
\end{figure}

Figure~\ref{fig:perclass_full} presents the same class-wise evidence graphically. The figure shows that DALight-3D maintains high specificity across all classes, while the principal weakness lies in sensitivity and overlap for ED. Figure~\ref{fig:confusion} further characterizes the error structure through a row-normalized confusion matrix with raw voxel counts. This experiment measures how voxels from each true class are redistributed among predicted classes. The matrix confirms that BG, NCR, and ET are comparatively well preserved, whereas ED exhibits the strongest confusion with neighboring categories.

\subsubsection{Calibration}

In addition to segmentation accuracy, we evaluate the reliability of the predicted probabilities through Expected Calibration Error (ECE)~\cite{guo2017calibration} using 15 bins. This experiment measures whether the predicted confidence is aligned with observed correctness rather than merely producing accurate masks. At the best retained validation checkpoint, DALight-3D attains an ECE of approximately 0.009 together with an overall voxel-wise accuracy of 0.974, indicating that its probabilities remain well behaved in addition to being discriminative.

\subsection{Discussion}
\label{sec:discussion}

The results indicate that the proposed design provides a strong accuracy--efficiency trade-off in this benchmark. DALight-3D attains the highest mean Dice in the primary comparison while using fewer parameters than every comparison model. From a methodological perspective, this pattern is important because it suggests that compact 3D segmentation performance need not depend solely on widening the backbone or increasing training complexity. Instead, the results support a more targeted design strategy in which computationally expensive parts of the encoder--decoder pipeline are replaced by modules that are cheaper but more selective in how they process spatial, contextual, and skip-level information.

The baseline comparison and ablation study together provide a coherent interpretation of where this gain likely comes from. SepConv reduces the cost of volumetric processing without collapsing representational quality, which is particularly important in 3D where kernel cost scales rapidly with channel width. CSA appears to add context where it matters most, namely in the deeper stages where semantic abstraction is strongest and the spatial cost of attention is lower. Identifier-conditioned normalization contributes a smaller but still meaningful gain, suggesting that lightweight feature modulation remains useful even when only proxy identifiers are available. SSFB provides an additional reconstruction benefit by making skip fusion more selective than direct concatenation. Taken together, these findings support the view that the observed improvement is the result of complementary design choices rather than a single dominant architectural trick.

The class-wise results further refine this interpretation. DALight-3D performs strongest on NCR and ET, whereas ED remains the most difficult region. This pattern is consistent with prior observations in BraTS-style segmentation, where edema is typically more diffuse, less structurally compact, and often less sharply bounded than enhancing or core tumor regions. The retained performance gap on ED therefore highlights a clinically plausible failure mode rather than an anomalous result. At the same time, the low ECE indicates that the model retains useful calibration properties, which is desirable for uncertainty-aware downstream interpretation even when segmentation quality varies by class.

Several limitations should be acknowledged. First, the current study is restricted to a single benchmark dataset; using the full BraTS-derived Task01\_BrainTumour benchmark strengthens coverage within that dataset, but it does not by itself establish cross-dataset generalization. Second, the conditioning signal used here is based on deterministic proxy buckets rather than curated scanner metadata, so the present results should not be interpreted as a direct multi-scanner robustness evaluation. Third, the reported model comparisons correspond to single completed runs, so run-to-run variability is not quantified here. Fourth, although the ablation models were trained for 50 total epochs, only the final 25 epochs were retained in the saved ablation histories, which limits curve-level analysis. Accordingly, the present study is best interpreted as a reproducible benchmark analysis of a compact 3D architecture rather than as a definitive robustness study across acquisition domains. Future work should therefore include experiments with curated scanner metadata, external or unseen-site cohorts, multiple runs per model, and fully retained ablation logs.
\vspace{-4pt}
\section{Conclusion}
\label{sec:conclusion}
\vspace{-1pt}
This paper introduced \textbf{DALight-3D}, a lightweight 3D segmentation architecture for multi-modal brain tumor MRI. The method revisits the encoder--decoder paradigm through four targeted modifications: separable 3D convolution for computational efficiency, identifier-conditioned normalization for feature modulation, cross-slice attention for contextual reasoning, and SSFB for adaptive multi-scale fusion.

On Medical Segmentation Decathlon Task01\_BrainTumour, DALight-3D achieves a mean Dice of 0.727 with 2.22M parameters, whereas Residual 3D U-Net achieves 0.710 Dice with 3.20M parameters. The ablation study further supports the contribution of all four proposed components, with the clearest penalties observed when SepConv or CSA is removed. Overall, these results position DALight-3D as a compact baseline for 3D brain tumor segmentation in a reproducible benchmark study. Future work will focus on multiple-run evaluation, systematic ablation with fully retained logs, and assessment with curated scanner metadata or external cohorts.

\section*{Declarations}

\subsection*{Availability of data and material}
The dataset analyzed in this study is publicly available from the Medical Segmentation Decathlon repository (Task01\_BrainTumour): \url{http://medicaldecathlon.com/}. The code, experiment configuration, and paper-release artifacts are publicly available at: \url{https://github.com/Dhruv27Mishra/DALight3D-paper-release}.

\subsection*{Competing interests}
The authors declare that they have no competing interests.

\subsection*{Funding}
No external funding was received for this work.

\subsection*{Authors' contributions}
Nand Kumar Mishra conceptualized the study, developed the methodology, and supervised the research. Dhruv Mishra implemented the models, conducted experiments, curated data processing, and prepared figures. Dr.\ Manu Pratap Singh provided technical guidance, validated the analysis, and reviewed the manuscript critically. All authors read and approved the final manuscript.

\subsection*{Acknowledgements}
The authors thank Dr.\ Bhimrao Ambedkar University and Shiv Nadar University for institutional support. The authors also acknowledge Google Colab for providing the computational environment used in this study.

\IfFileExists{Main.bbl}{
}{\bibliography{refs}}

\end{document}